\documentclass[10pt, conference]{IEEEtran}
\IEEEoverridecommandlockouts
\usepackage{cite}
\usepackage{amsmath, amssymb, amsfonts}
\usepackage{algorithmic}
\usepackage{graphicx}
\usepackage{textcomp}
\usepackage{xcolor}
\usepackage{booktabs}
\usepackage{multirow}
\usepackage{array}
\usepackage{makecell}
\usepackage{longtable}
\usepackage{subcaption}
\usepackage{tikz}
\usetikzlibrary{calc, positioning, decorations.pathreplacing, arrows.meta}
\usepackage{pgfplots}
\pgfplotsset{compat=1.18}
\usepackage{placeins}
\usepackage{dblfloatfix}
\usepackage{url}
\usepackage[hidelinks]{hyperref}
\usepackage{tabularx}
\usepackage{balance}   

\def\BibTeX{{\rm B\kern-.05em{\sc i\kern-.025em b}\kern-.08em T\kern-.1667em\lower.7ex\hbox{E}\kern-.125emX}}

\graphicspath{{figures/}}

\begin{document}
    \newcommand{\TODO}[1]{\textcolor{red}{[#1]}}
\newcommand{\NumEncoders}{64}
\newcommand{\NumGroups}{8}
\newcommand{\NumEpochs}{100}
\newcommand{\NumDegenerate}{25}
\newcommand{\MedianEncParams}{11.7}
\newcommand{\MinEncParams}{0.2}
\newcommand{\MaxEncParams}{100.3}
\newcommand{\TrainW}{512}
\newcommand{\TrainH}{288}
\newcommand{\TrainBS}{12}
\newcommand{\BestEncoder}{internimage\_t}
\newcommand{\BestScore}{0.688}
\newcommand{\BestEncDelta}{0.725}
\newcommand{\BestEncMIoU}{0.650}
\newcommand{\BestEncEPE}{10.85}
\newcommand{\BestSegEncoder}{densenet121}
\newcommand{\BestMIoU}{0.654}
\newcommand{\BestDepthEncoder}{internimage\_t}
\newcommand{\BestDeltaOne}{0.725}
\newcommand{\BestAbsRelEncoder}{swin\_tiny\_w4}
\newcommand{\BestAbsRel}{0.358}
\newcommand{\BestEPEEncoder}{internimage\_t}
\newcommand{\BestEPE}{10.85}
\newcommand{\BestGroup}{CNN-Transformer Hybrid}
\newcommand{\BestGroupLetter}{G}
\newcommand{\BestGroupScore}{0.600}
\newcommand{\WorstGroup}{Plain Transformer}
\newcommand{\WorstGroupLetter}{E}
\newcommand{\WorstGroupScore}{0.329}
\newcommand{\RhoSegDepth}{0.89}
\newcommand{\RhoSegDepthN}{64}
\newcommand{\SmallStrongEncoder}{edgenext\_xxs}
\newcommand{\SmallStrongParams}{1.2}
\newcommand{\WeakEncoder}{resmlp\_12}
\newcommand{\WeakEncoderDelta}{0.032}
\newcommand{\WeakEncoderEPE}{44.64}
\newcommand{\DegenEncoder}{resnext50\_16x4d}
\newcommand{\DegenGroup}{A}
\newcommand{\DegenParams}{12.9}
\newcommand{\DegenTreeIoU}{0.929}
\newcommand{\DegenIoUBg}{0.012}
\newcommand{\DegenBFOne}{0.000}
\newcommand{\DegenMIoU}{0.471}
\newcommand{\BestEncIoUBg}{0.357}
\newcommand{\BestEncIoUTree}{0.943}
\newcommand{\BestEncBFOne}{0.419}
\newcommand{\BestEncMedAE}{0.20}
\newcommand{\BestEncPninetyAE}{2.45}
\newcommand{\BestEncMAE}{1.39}
\newcommand{\BestEncBadTwo}{0.391}
\newcommand{\BestEncAbsRel}{0.786}
\newcommand{\LiveMIoULo}{0.314}
\newcommand{\LiveMIoUHi}{0.654}
\newcommand{\LiveDeltaLo}{0.032}
\newcommand{\LiveDeltaHi}{0.725}
\newcommand{\NumLive}{39}
\newcommand{\NumPlainT}{4}
\newcommand{\NumPlainTDegen}{3}
\newcommand{\PlainTException}{deit\_tiny\_patch16}
\newcommand{\PlainTExceptionParams}{6.4}
\newcommand{\PlainTExceptionRank}{23}
\newcommand{\NumInBudget}{22}
\newcommand{\PctInBudget}{34}
\newcommand{\NumScenesTest}{5}
\newcommand{\NumFramesTest}{240}
\newcommand{\BootN}{2000}
\newcommand{\BootRho}{0.994}
\newcommand{\BootRhoLo}{0.962}
\newcommand{\BootTopOne}{100}


\title{What Does the Encoder Actually Decide? A Controlled Comparison of
\NumEncoders{} Vision Backbones on Joint Tree Segmentation and Stereo Depth}

\author{\IEEEauthorblockN{Yida Lin, Bing Xue, Mengjie Zhang} \IEEEauthorblockA{\small \textit{Centre for Data Science and Artificial Intelligence} \\ \textit{Victoria University of Wellington, Wellington, New Zealand}\\ linyida\texttt{@}myvuw.ac.nz, bing.xue\texttt{@}vuw.ac.nz, mengjie.zhang\texttt{@}vuw.ac.nz}
    \and \IEEEauthorblockN{Sam Schofield, Richard Green} \IEEEauthorblockA{\small \textit{Department of Computer Science and Software Engineering} \\ \textit{University of Canterbury, Canterbury, New Zealand}\\ sam.schofield\texttt{@}canterbury.ac.nz, richard.green\texttt{@}canterbury.ac.nz}}

\maketitle
\vspace{-1.4em}

\begin{abstract}
    A robot that prunes trees needs two facts about the same pixel: whether it
    belongs to a tree, and how far away it is. Both are normally obtained by
    attaching a task head to a vision backbone, and the backbone is chosen by
    reputation rather than by measurement. We ask a narrow but answerable
    question: with the dataset, the decoders, the losses, the schedule and the
    evaluation all held fixed, how much does the choice of encoder actually
    change joint semantic segmentation and stereo depth on thin vegetation? We
    build a hard parameter-sharing network in which one shared encoder feeds a
    segmentation branch and a stereo branch, and in which \emph{only} the
    encoder is swapped --- a one-string change that retunes nothing downstream.
    We evaluate \NumEncoders{} encoders, one canonical representative per
    architecture family across \NumGroups{} groups (classical, lightweight and
    modern CNNs, plain and hierarchical transformers, hybrids, MLP-mixers and
    state-space models), each at the preset nearest a common $\sim$25\,M budget
    and trained from scratch so that a pretrained checkpoint is not the
    confound; where a family publishes nothing near that budget, capacity is
    analysed explicitly rather than assumed away.
    Evaluation is deliberately unforgiving: depth is scored on tree
    pixels only, and segmentation is scored with boundary F1 and background IoU
    that a ``label-everything-tree'' collapse cannot win. Three findings stand
    out. First, the strongest encoders are convolutional and hybrid, not
    transformer: \BestEncoder{} leads with \BestEncMIoU{} segmentation mIoU and
    \BestEncDelta{} depth $\delta_1$, while \NumPlainTDegen{} of the
    \NumPlainT{} plain vision transformers collapse when trained from scratch.
    Second, parameters do not predict quality --- \SmallStrongEncoder{}, at only
    \SmallStrongParams{}\,M encoder parameters, outranks models nearly two
    orders of magnitude larger. Third,
    the segmentation and depth rankings agree strongly (Spearman
    $\rho=\RhoSegDepth{}$), so the shared encoder faces no genuine task conflict
    and a single backbone serves both. Along
    the way, \NumDegenerate{} of \NumEncoders{} encoders collapse to a
    degenerate all-tree segmentation that region IoU hides but boundary F1
    exposes --- a reminder that the metric, not only the backbone, decides what
    a comparison can see.
\end{abstract}


\vspace{-0.8em}
\section{Introduction}

A robot that prunes trees must answer two questions about the same pixel. The
first is one of membership: is this pixel part of a tree, or part of the
background behind it? The second is one of geometry: how far away is it?
Neither answer is useful alone. Perfect segmentation without depth cannot guide
a cutter; perfect depth without a tree mask cannot tell the cutter where the
wood is.

Both tasks are, in practice, built the same way: take a vision backbone, attach
a decoder, and train. The backbone holds most of the parameters and most of the
compute, and it is usually chosen by reputation --- whatever performed well on
ImageNet classification, or whatever the previous paper in the area used. That
inheritance is rarely questioned, and it is not obviously safe. Tree canopies
are not ImageNet: the objects are thin, self-occluding, repetitive and
semi-transparent, and the discriminative signal lives in high-frequency
boundaries rather than in object-level texture.

This paper asks a deliberately narrow question, because the narrow version is
the one that can be answered: \emph{with everything else held fixed, how much
does the encoder decide?} We are not proposing a new architecture. We are
measuring how much the architecture choice is worth on a task whose conventional
wisdom was formed somewhere else.

Answering it requires a setup in which the encoder is genuinely the only thing
that changes, and an evaluation that cannot be won by a shortcut: trees fill
\mbox{75--91\%} of every tree50 frame, so region IoU is nearly saturated before
a model has learned anything. We therefore score depth on tree pixels only and
segmentation with boundary F1 and background IoU, which a
``label-everything-tree'' collapse cannot win. Our contributions are:

\begin{itemize}
    \item \textbf{A two-task benchmark in which the encoder is the only free
    variable.} A hard parameter-sharing network feeds one shared encoder into a
    segmentation branch and a stereo branch. Every encoder exposes the same
    five-scale feature contract, so swapping it is a one-string change; the
    decoders, losses, schedule and evaluation never move.

    \item \textbf{\NumEncoders{} encoders, one per family, from scratch.} One
    canonical representative per architecture family across \NumGroups{}
    groups, taken at the preset nearest a common budget with ablation-only
    variants excluded, and trained without pretrained weights so that a
    checkpoint is not the confound. Where families publish nothing near that
    budget, capacity is measured rather than assumed away.

    \item \textbf{An evaluation designed to resist the obvious cheat.} Depth on
    tree pixels only; segmentation reported with boundary F1 and background IoU
    alongside region IoU, so a degenerate all-tree solution is visible rather
    than rewarded.

    \item \textbf{A measured answer.} Convolutional and hybrid encoders lead
    while most plain transformers collapse from scratch; parameter count does
    not predict quality; and the segmentation and depth orderings are strongly
    correlated, so one shared backbone suffices for both tasks.
\end{itemize}

\vspace{-0.3em}
\section{Related Work}

\subsection{Perception for Tree and Orchard Robotics}

Robots that work on woody plants --- harvesting, thinning, pruning, forwarding
--- have converged on the same perception requirement: a per-pixel decision
about what is plant and where it is in
space~\cite{bac2014harvesting, zahid2021pruner}. Field pruning systems report
that perception, not manipulation, is the binding
constraint~\cite{you2022pruning}, and the first autonomous forestry machines
make the same observation at a larger scale~\cite{lahera2024forestry}. What
these systems need from a network is exactly the pair we study: a tree mask
that follows the true silhouette, and a depth estimate on the pixels inside it.

Two properties of the domain make it a poor match for the assumptions behind
ImageNet-era backbone design. First, the targets are \emph{thin}: branches and
twigs are a few pixels wide, so the signal lives in high-frequency boundaries
rather than object-level texture --- the same difficulty that motivated
dedicated treatment of wires and cables in aerial
robotics~\cite{madaan2017wire}. Second, exact labels on thin structures are
effectively unobtainable by hand, which is why rendered data has become a
standard instrument for agricultural
segmentation~\cite{barth2018synthesis} and is what we use here. What is
missing from this literature is a controlled measurement of how much the
\emph{backbone} contributes, which is the gap this paper addresses.

\subsection{Vision Backbones}

The encoder families we compare correspond to the main lines of backbone
research. Classical convolutional networks established depth, residual
connections and cardinality as the levers of representational
power~\cite{he2016resnet, simonyan2015vgg, huang2017densenet, xie2017resnext}.
Efficiency-driven designs traded dense convolution for depthwise-separable and
shuffled operations to reach mobile budgets~\cite{howard2019mobilenetv3,
ma2018shufflenetv2, ding2021repvgg}. Vision transformers replaced the inductive
bias of convolution with global attention~\cite{dosovitskiy2021vit,
touvron2021deit}, and hierarchical variants reintroduced locality and
multi-scale structure to make attention affordable for dense
prediction~\cite{liu2021swin, wang2021pvt}. Modern convolutional networks then
showed that much of the transformer advantage could be recovered by
convolutional designs with large kernels and modernised training
recipes~\cite{liu2022convnext, ding2022replknet}. Hybrids interleave the two
families~\cite{dai2021coatnet, yu2022metaformer}, attention-free MLP designs
question whether attention is required at all~\cite{tolstikhin2021mlpmixer,
trockman2023convmixer}, and state-space models offer linear-complexity sequence
mixing as a third alternative~\cite{liu2024vmamba, zhu2024vim}.

These families are usually compared on ImageNet classification, occasionally on
COCO detection or ADE20K segmentation. We are not aware of a controlled
comparison on a task that couples dense semantic segmentation with stereo
geometry on thin structures, which is the regime this paper targets.

\subsection{Multi-Task Learning}

Sharing one encoder between related tasks is the standard way to amortise
computation, and hard parameter sharing --- one shared trunk, independent
task-specific heads --- remains the reference
formulation~\cite{caruana1997multitask, kendall2018multitask}. Richer schemes
add cross-task connections, cascades or attention between
branches~\cite{misra2016crossstitch, vandenhende2021mtlsurvey}. We deliberately
use the plain hard-sharing form: any cross-task pathway would give the encoder a
second route to influence the result and would blur exactly the attribution
this paper is trying to make.

\subsection{Stereo Matching and Segmentation}

Our stereo branch follows the cost-volume lineage: features are correlated
across a disparity range, aggregated with 3D convolutions, and reduced by
soft-argmin~\cite{kendall2017gcnet, chang2018psmnet, guo2019gwcnet}. The
segmentation branch is a standard U-Net encoder--decoder~\cite{ronneberger2015unet}
that fuses the five encoder scales into a per-pixel tree/background labelling.
Both branches are held fixed across all \NumEncoders{} encoders; the only thing
that changes between runs is the shared trunk that feeds them.

\vspace{-0.3em}
\section{Benchmark Design}
\label{sec:benchmark}

\subsection{Data}

We use the tree50 corpus: 50 rendered forest scenes, each captured from 48
viewpoints, giving 2{,}400 rectified stereo pairs at $1920\times1080$ with
per-pixel depth and tree masks. Splits are \emph{scene-exclusive} --- scenes
1--40 train, 41--45 validate, 46--50 test --- so no tree is ever seen from a
different angle at test time. Because the renderer exports geometry rather than
hand annotation, the labels are exact on precisely the thin structures that
manual labelling cannot reach.

We state the size of the test set in the units that matter: it is
\NumFramesTest{} frames but only \NumScenesTest{} \emph{independent scenes},
since the 48 views of one scene show the same trees. Everything reported below
is therefore an average over \NumScenesTest{} trees, and
Section~\ref{subsec:stability} quantifies what that costs in confidence.

Two measured properties of the data shaped the design, and each is enforced by
an assertion in the codebase rather than left as a convention. First, disparity
is an exact function of depth, $d = f_x B / z$, agreeing with the rendered
output to $0.0$\,px on average; depth and disparity are therefore the same
signal in two units, so the network predicts disparity once and converts.
Second, trees occupy \mbox{75--91\%} of the pixels. This is why region IoU is a
weak metric here and why depth is scored on tree pixels only
(Section~\ref{sec:eval}).

\subsection{Network}

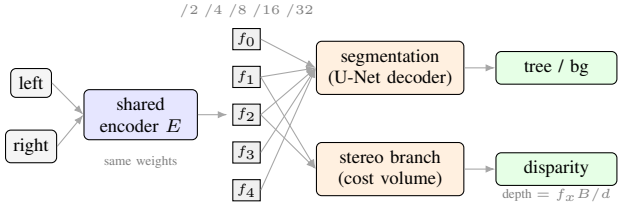
\begin{figure}[t]
    \centering
    \footnotesize
    \begin{tikzpicture}[
        box/.style={draw, rounded corners=2pt, align=center, inner sep=3pt,
                    font=\scriptsize},
        enc/.style={box, fill=blue!10, minimum width=1.5cm, minimum height=0.6cm},
        feat/.style={draw, fill=black!6, minimum width=0.30cm, inner sep=1.5pt,
                     font=\tiny},
        head/.style={box, fill=orange!12, minimum width=1.95cm, minimum height=0.5cm},
        outbox/.style={box, fill=green!10, minimum width=1.6cm, minimum height=0.42cm},
        ar/.style={-{Latex[length=1.3mm]}, gray!70, line width=0.4pt},
    ]
        \node[box, fill=black!4] (l) at (0, 0.42) {left};
        \node[box, fill=black!4] (r) at (0, -0.42) {right};
        \node[enc] (e) at (1.45, 0) {shared\\encoder $E$};
        \draw[ar] (l.east) -- (e.west);
        \draw[ar] (r.east) -- (e.west);
        \node[font=\tiny, gray] at (1.45, -0.62) {same weights};

        \foreach \i/\lab/\y in {0/{$f_0$}/1.0, 1/{$f_1$}/0.5, 2/{$f_2$}/0.0,
                                3/{$f_3$}/-0.5, 4/{$f_4$}/-1.0}{
            \node[feat] (f\i) at (2.85, \y) {\lab};
        }
        \node[font=\tiny, gray] at (2.85, 1.38) {$/2\;/4\;/8\;/16\;/32$};
        \draw[ar] (e.east) -- (2.6, 0);

        \node[head] (seg) at (4.75, 0.62) {segmentation\\(U-Net decoder)};
        \node[head] (st)  at (4.75, -0.72) {stereo branch\\(cost volume)};
        \foreach \i in {0,1,2,3,4}{ \draw[ar] (f\i.east) -- (seg.west); }
        \draw[ar] (f1.east) -- (st.west);
        \draw[ar] (f2.east) -- (st.west);

        \node[outbox] (o1) at (6.95, 0.62) {tree / bg};
        \node[outbox] (o4) at (6.95, -0.72) {disparity};
        \draw[ar] (seg.east) -- (o1.west);
        \draw[ar] (st.east) -- (o4.west);
        \node[font=\tiny, gray] at (6.95, -1.06) {depth $=f_xB/d$};
    \end{tikzpicture}
    \caption{Hard parameter sharing. One encoder, applied to both views with
    shared weights, produces five feature scales; a segmentation branch and a
    stereo branch consume them independently, with no connection between the
    branches. Only the encoder is swapped --- everything to the right of
    $f_0\ldots f_4$ is identical across all \NumEncoders{} runs.}
    \label{fig:arch}
\end{figure}

The left and right images pass through the \emph{same} encoder weights,
producing five feature maps at strides $2,4,8,16,32$ (Fig.~\ref{fig:arch}). The
segmentation branch is a U-Net decoder over all five scales, emitting a
per-pixel tree/background labelling. The stereo branch builds a group-wise
correlation cost volume from the left and right stride-4 features, aggregates it
with 3D convolutions, reduces it by soft-argmin, and refines the result
residually; depth follows from disparity by $z=f_xB/d$. Beyond the shared
encoder there is no connection between the branches.\footnote{The segmentation
decoder also carries centre and offset heads for an instance-grouping extension
outside the scope of this paper. They are neither analysed nor reported, but
they are present and supervised in \emph{every} run, so they shift the absolute
numbers for all encoders alike rather than favouring any one of them.}

\subsection{The Five-Scale Contract}

The comparison is only meaningful if the encoder is genuinely interchangeable.
Every encoder implements one method, \texttt{forward\_features}, returning five
maps whose $i$-th spatial size is exactly $\lceil H/2^{i+1} \rceil$. The
contract fixes \emph{ratios}, not sizes, so the same encoder serves any input
resolution; only the channel counts differ. Inputs are reflection-padded to a
multiple of 32 and each level is cropped back, so padding never leaks into the
feature sizes.

\subsection{Encoder Selection}

Two levels of organisation are used throughout and are not interchangeable. A
\emph{family} is one published architecture line (ResNeXt, Swin, ConvNeXt,
\ldots); a \emph{group} is the broad design paradigm it belongs to, of which
there are \NumGroups{}, labelled A--I. We select \emph{one representative per
family}, giving \NumEncoders{} encoders spread over the \NumGroups{} groups.

The selection applies three rules, in order.
First, \textbf{exclude ablation-only variants}: several families ship control
conditions alongside their proposal --- a large-kernel network with the kernels
reduced to $3\times3$, a state-space model with the state-space mixing removed
--- and representing a family by its own control would understate it. Second,
\textbf{target the parameter budget}: among the remainder we take the preset
closest to a common $\sim$25\,M encoder budget, the tier at which most families
publish their reference model. Third, \textbf{prefer canonical naming} when the
first two rules tie. A small hand-curated override fixes the cases where the
automatic rule still picks a non-canonical variant.

We state plainly that this \emph{targets} a budget rather than matching one.
Families do not publish on a common grid: only \NumInBudget{} of
\NumEncoders{} encoders (\PctInBudget{}\%) land within $\pm$10\,M of the
target, the median is \MedianEncParams{}\,M, and the range is
\MinEncParams{}--\MaxEncParams{}\,M --- four families whose smallest published
variant is a $\sim$100\,M base model are represented at that size. Rather than
discard the families that do not fit, we keep them and treat capacity as a
measured covariate: Fig.~\ref{fig:pareto} plots quality directly against
encoder size, and the result --- that the largest models are among the worst
--- is what licenses reading the rest of the comparison as being about
architecture rather than about scale.

\vspace{-0.3em}
\section{Evaluation Protocol}
\label{sec:eval}

All metrics are computed on the left image of the test split.

\subsection{Segmentation}

We report tree-versus-background mean IoU (mIoU), the individual background and
tree IoUs, boundary F1 at a 2\,px tolerance (BF1), and pixel accuracy. The
metric we consider most diagnostic is not region IoU but \textbf{boundary F1}
together with \textbf{background IoU}. Because trees fill most of the frame, a
model that simply labels every pixel ``tree'' already scores $\sim$0.93 tree IoU
while contributing nothing: its background IoU is near zero and its boundary F1
collapses. Region IoU hides this failure; the boundary and background terms
expose it. Since the downstream task needs the tree \emph{silhouette}, the
boundary term is the one that matters.

\subsection{Depth}

Depth is evaluated \textbf{on ground-truth tree pixels only}. Sky and ground are
geometrically trivial and would dilute the metric with the easy majority. We
report AbsRel, RMSE, SILog and $\delta<1.25^k$, plus disparity EPE and
bad-$\tau$ rates on the same pixel set.

One property of this pixel set deserves stating, because it makes the mean
error look worse than the prediction is. Tree pixels include distant twigs at
the far end of the disparity range, where a sub-pixel disparity error maps to a
metric depth error of several metres. The absolute error is therefore strongly
right-skewed, and mean-based scores are set by that tail rather than by the
typical pixel: the best encoder has \BestEncAbsRel{} AbsRel and
\BestEncMAE{}\,m mean absolute error, but a \emph{median} absolute error of
\BestEncMedAE{}\,m, with \BestEncPninetyAE{}\,m at the 90th percentile. We
report the mean-based metrics because they are the convention and keep this
work comparable, but read them as tail statistics; the median and the
percentiles describe what a controller would see on most of the branch it is
actually reaching for.

Two secondary views are reported because
they answer questions the primary one hides: \emph{per-tree} averaging weights
each tree equally rather than each pixel, so a model that has learned only the
large foreground tree separates from one that also handles the distant small
ones; and depth restricted to the \emph{predicted} tree mask measures
end-to-end usability, since a segmentation error propagates into the depth that
is actually consumed.

\subsection{Efficiency}

We report end-to-end FPS, peak memory, encoder parameters and GFLOPs, measured
on the whole two-branch network. Since the decoders are byte-identical across
runs, differences between rows are attributable to the encoder even though the
figures themselves are whole-system costs.

\vspace{-0.3em}
\section{Experimental Setup}
\label{sec:setup}

All encoders are trained \textbf{from scratch} --- no pretrained weights --- so
that the comparison measures architecture rather than the availability and
quality of a checkpoint. This lowers absolute accuracy and is a deliberate
trade; it also means the ranking here need not match one obtained with
pretrained initialisation, a point we return to in the limitations.

Every encoder receives an identical schedule: AdamW at learning rate
$3\times10^{-4}$, weight decay $10^{-2}$, cosine decay after linear warmup,
mixed precision, gradient clipping at norm $1.0$, and \NumEpochs{} epochs at
$\TrainW\times\TrainH$ with batch size \TrainBS{}. Normalisation layers and
biases are excluded from weight decay. The checkpoint is selected by lowest
total validation loss rather than by any single task metric, since selecting on
one metric would quietly turn the run into single-task model selection. The
segmentation branch is supervised by cross-entropy and the stereo branch by
smooth-$L_1$ on disparity at both the coarse and refined stages. The cost volume
is built at stride 4 over a disparity range that covers the native maximum, and
the segmentation decoder uses channel widths $(128, 96, 64, 48)$.

Every resolution-dependent constant --- chiefly the disparity search range and
the boundary-F1 tolerance --- is defined once at native resolution and derived
for any other resolution by a single function, so that supervision and
evaluation cannot silently disagree. Encoders are trained in isolated
subprocesses so that a memory failure in one cannot corrupt the allocator state
of the next. Where an encoder does not fit at the nominal batch size, the
micro-batch is reduced and gradient accumulation is raised to hold the effective
batch constant; holding the effective batch fixed matters because batch size
changes normalisation statistics and gradient noise, and an encoder penalised
for a memory property rather than for its representational quality would defeat
the purpose of the comparison. State-space and global-attention encoders are the
systematic pressure point here, since their activation memory far exceeds what
their parameter count suggests; of the full selection, \NumEncoders{} encoders
completed training and are reported below.

\vspace{-0.3em}
\section{Results}
\label{sec:results}

\begin{table*}[t]
\centering
\caption{Per-encoder results on the tree50 test split, sorted by the composite
score (Section~\ref{sec:results}); the ranking continues in the right-hand
block. Segmentation is scored on the left image, depth on ground-truth tree
pixels only. P$_{\text{e}}$ is encoder parameters (M). Best in each column
is bold; $\uparrow$/$\downarrow$ mark higher/lower is better. mIoU is the
tree-vs-background mean IoU; IoU$_{\text{bg}}$ and BF1 are the
discriminative components a ``label-everything-tree'' collapse cannot win, and
the run of near-zero entries in those two columns is the block of collapsed
encoders discussed in Section~\ref{sec:results}.}
\label{tab:main}
\scriptsize
\setlength{\tabcolsep}{3pt}
\renewcommand{\arraystretch}{0.95}
\begin{tabular}{@{}rllrrrrrr@{\hspace{1.6em}}rllrrrrrr@{}}
\toprule
\# & Encoder & Grp & P$_{\text{e}}$ & mIoU & IoU$_{\text{bg}}$ & BF1 & AbsRel & $\delta_1$ & \# & Encoder & Grp & P$_{\text{e}}$ & mIoU & IoU$_{\text{bg}}$ & BF1 & AbsRel & $\delta_1$ \\
 & & & (M) & $\uparrow$ & $\uparrow$ & $\uparrow$ & $\downarrow$ & $\uparrow$ &  & & & (M) & $\uparrow$ & $\uparrow$ & $\uparrow$ & $\downarrow$ & $\uparrow$ \\
\midrule
1 & internimage\_t & D & 28.8 & 0.650 & 0.357 & 0.419 & 0.786 & \textbf{0.725} & 33 & convmixer\_512\_12 & H & 9.9 & 0.577 & 0.215 & 0.114 & 0.562 & 0.525 \\
2 & wrn\_40\_4 & A & 16.1 & 0.649 & 0.358 & 0.444 & 0.465 & 0.699 & 34 & zfnet\_slim & A & 6.4 & 0.590 & 0.237 & 0.231 & 1.750 & 0.499 \\
3 & swin\_tiny\_w4 & F & 27.5 & 0.633 & 0.319 & 0.402 & \textbf{0.358} & 0.701 & 35 & resnet18 & A & 11.2 & 0.596 & 0.281 & 0.467 & 0.732 & 0.491 \\
4 & xception\_mobile\_order & A & 20.8 & 0.634 & 0.321 & 0.389 & 0.843 & 0.687 & 36 & cyclemlp\_b1 & H & 14.7 & 0.586 & 0.231 & 0.164 & 0.782 & 0.490 \\
5 & res2next50 & A & 22.6 & 0.635 & 0.325 & 0.369 & 0.545 & 0.680 & 37 & crossformer\_tiny & F & 27.4 & 0.569 & 0.199 & 0.103 & 0.667 & 0.486 \\
6 & edgenext\_xxs & G & 1.2 & 0.626 & 0.309 & 0.411 & 0.797 & 0.686 & 38 & mvitv1\_tiny & F & 23.4 & 0.517 & 0.100 & 0.015 & 0.593 & 0.524 \\
7 & biformer\_tiny & F & 12.6 & 0.597 & 0.252 & 0.275 & 1.135 & 0.702 & 39 & hornet\_tiny & D & 21.9 & 0.471 & 0.013 & 0.005 & 0.654 & 0.568 \\
8 & convnext\_atto & D & 3.4 & 0.606 & 0.268 & 0.325 & 0.564 & 0.691 & 40 & hiremlp\_tiny & H & 32.0 & 0.466 & 0.004 & 0.001 & 0.985 & 0.525 \\
9 & coat\_lite\_tiny & G & 6.1 & 0.615 & 0.288 & 0.377 & 1.295 & 0.667 & 41 & moganet\_xtiny & D & 2.8 & 0.483 & 0.037 & 0.011 & 0.743 & 0.505 \\
10 & hrnet\_w18\_small & A & 3.9 & 0.619 & 0.295 & 0.374 & 0.516 & 0.651 & 42 & efficientformerv2\_s0 & G & 2.4 & 0.480 & 0.030 & 0.003 & 0.594 & 0.505 \\
11 & inception\_v3\_small & A & 4.5 & 0.615 & 0.285 & 0.305 & 0.411 & 0.655 & 43 & mobilenetv4\_conv\_s & C & 0.8 & 0.482 & 0.034 & 0.004 & 1.044 & 0.486 \\
12 & densenet121 & A & 7.0 & \textbf{0.654} & \textbf{0.367} & \textbf{0.489} & 0.692 & 0.610 & 44 & fasternet\_t0 & C & 2.6 & 0.485 & 0.040 & 0.012 & 1.298 & 0.469 \\
13 & unireplknet\_a & D & 4.0 & 0.598 & 0.254 & 0.223 & 1.717 & 0.646 & 45 & focalnet\_tiny\_srf & D & 27.7 & 0.472 & 0.015 & 0.003 & 0.991 & 0.477 \\
14 & ceit\_tiny\_nolca & G & 6.5 & 0.629 & 0.311 & 0.375 & 1.413 & 0.612 & 46 & mambavision\_t & I & 31.2 & 0.464 & 0.000 & 0.000 & 0.755 & 0.482 \\
15 & levit\_128s & G & 5.4 & 0.622 & 0.300 & 0.379 & 1.256 & 0.608 & 47 & sequencer2d\_s & H & 28.3 & 0.464 & 0.000 & 0.000 & 1.399 & 0.461 \\
16 & container\_light & G & 22.9 & 0.581 & 0.219 & 0.181 & 1.338 & 0.645 & 48 & se\_resnet18 & A & 11.3 & 0.472 & 0.015 & 0.002 & 0.696 & 0.434 \\
17 & slak\_tiny & D & 49.1 & 0.597 & 0.256 & 0.267 & 0.402 & 0.626 & 49 & vgg16\_half & A & 3.7 & 0.465 & 0.002 & 0.000 & 0.869 & 0.435 \\
18 & cmt\_tiny & G & 8.7 & 0.579 & 0.217 & 0.166 & 0.437 & 0.626 & 50 & resnext50\_16x4d & A & 12.9 & 0.471 & 0.012 & 0.000 & 1.370 & 0.407 \\
19 & hiera\_tiny & F & 27.1 & 0.602 & 0.260 & 0.262 & 0.434 & 0.594 & 51 & efficientvit\_mit\_b0 & G & 0.7 & 0.464 & 0.000 & 0.001 & 0.735 & 0.402 \\
20 & wavemlp\_t\_dw & H & 13.1 & 0.631 & 0.317 & 0.388 & 0.688 & 0.562 & 52 & micronet\_m0 & C & 0.2 & 0.464 & 0.000 & 0.000 & 0.603 & 0.379 \\
21 & poolformer\_s12 & G & 11.4 & 0.603 & 0.262 & 0.248 & 1.159 & 0.591 & 53 & pvtv2\_b0\_li & F & 3.1 & 0.464 & 0.000 & 0.000 & 0.789 & 0.376 \\
22 & davit\_tiny & F & 27.6 & 0.590 & 0.237 & 0.200 & 0.647 & 0.603 & 54 & flatten\_tiny & I & 48.2 & 0.464 & 0.000 & 0.000 & 4.819 & 0.375 \\
23 & deit\_tiny\_patch16 & E & 6.4 & 0.574 & 0.208 & 0.105 & 0.903 & 0.603 & 55 & mobilenetv1\_025 & C & 0.2 & 0.465 & 0.001 & 0.002 & 1.132 & 0.370 \\
24 & fastvit\_t8 & G & 3.1 & 0.605 & 0.269 & 0.265 & 0.555 & 0.568 & 56 & cswin\_tiny & F & 21.8 & 0.464 & 0.000 & 0.000 & 3.506 & 0.356 \\
25 & lenet5\_classic & A & 0.7 & 0.596 & 0.248 & 0.255 & 1.089 & 0.572 & 57 & alexnet & A & 2.9 & 0.464 & 0.000 & 0.000 & 4.692 & 0.292 \\
26 & conformer\_tiny & G & 24.6 & 0.595 & 0.251 & 0.240 & 0.547 & 0.558 & 58 & dpn68 & A & 11.8 & 0.464 & 0.000 & 0.000 & 3.201 & 0.292 \\
27 & coatnet\_0\_cccc & G & 16.0 & 0.595 & 0.252 & 0.320 & 0.621 & 0.556 & 59 & beit\_base\_patch16 & E & 99.7 & 0.467 & 0.006 & 0.004 & 4.283 & 0.227 \\
28 & replknet\_31\_tiny & D & 11.5 & 0.572 & 0.204 & 0.143 & 0.848 & 0.566 & 60 & cait\_xxs24 & E & 12.6 & 0.464 & 0.000 & 0.000 & 8.527 & 0.159 \\
29 & ghostnet\_0\_5 & C & 0.7 & 0.567 & 0.194 & 0.163 & 0.559 & 0.560 & 61 & mixer\_s16 & H & 24.3 & 0.418 & 0.102 & 0.187 & 2.815 & 0.154 \\
30 & regnetx\_002 & A & 2.3 & 0.543 & 0.149 & 0.125 & 0.507 & 0.564 & 62 & clip\_vit\_base\_patch16 & E & 100.3 & 0.464 & 0.000 & 0.000 & 1.485 & 0.107 \\
31 & maxvit\_tiny\_w4 & F & 28.6 & 0.489 & 0.046 & 0.003 & 0.567 & 0.615 & 63 & gmlp\_tiny & H & 6.1 & 0.314 & 0.108 & 0.132 & 14.940 & 0.103 \\
32 & asmlp\_tiny & H & 27.5 & 0.544 & 0.151 & 0.097 & 0.575 & 0.559 & 64 & resmlp\_12 & H & 18.5 & 0.365 & 0.056 & 0.092 & 122.315 & 0.032 \\
\bottomrule
\end{tabular}
\end{table*}

We order encoders by a single \textbf{composite score}, the mean of semantic
mIoU and depth $\delta_1$ (both higher-is-better and roughly in $[0,1]$). The
score is a convenience for ranking, not a proposed metric; every underlying
number is in Table~\ref{tab:main}, and the two tasks are also examined
separately below.

\subsection{Overall Comparison}

Table~\ref{tab:main} reports all \NumEncoders{} encoders in rank order. The
strongest is \BestEncoder{} (Modern CNN), which combines
\BestEncMIoU{} segmentation mIoU with the best depth of the whole study
(\BestEncDelta{} $\delta_1$, \BestEncEPE{}\,px EPE), for a composite of
\BestScore{}. It is followed by a classical wide residual network and a
hierarchical transformer, and the top of the ranking is dominated by
convolutional and hybrid designs. The best segmentation in isolation belongs to
\BestSegEncoder{} (\BestMIoU{} mIoU, and the highest boundary F1), and the best
disparity error to \BestEncoder{}; the lowest AbsRel is \BestAbsRelEncoder{} at
\BestAbsRel{}. No plain vision transformer appears near the top:
\NumPlainTDegen{} of the \NumPlainT{} collapse outright, and the exception,
\PlainTException{}, only reaches rank \PlainTExceptionRank{}
(Section~\ref{subsec:groups}).

The comparison also exposes a failure the region metric would have hidden.
\NumDegenerate{} of \NumEncoders{} encoders collapse to a degenerate solution
that labels almost everything ``tree'': their tree IoU stays near $0.93$, yet
their background IoU and boundary F1 fall to near zero and their mIoU drops to
$\sim$0.46. Region IoU alone would have ranked these models as competent; the
boundary and background terms correctly place them at the bottom. This is the
concrete payoff of the metric choice in Section~\ref{sec:eval}, and
Section~\ref{subsec:qual} shows what the collapse looks like on the image.

\subsection{By Architecture Family}
\label{subsec:groups}

\begin{table}[t]
\centering
\caption{Median performance per architecture-family group (median, not mean, so
one collapsed encoder cannot dominate). Score is the composite of
Section~\ref{sec:results}. Groups are ordered A--I.}
\label{tab:group}
\scriptsize
\setlength{\tabcolsep}{3pt}
\renewcommand{\arraystretch}{0.95}
\begin{tabular}{@{}llrrrrrrr@{}}
\toprule
& Group & $n$ & mIoU$\uparrow$ & BF1$\uparrow$ & AbsRel$\downarrow$ & $\delta_1\uparrow$ & EPE$\downarrow$ & Score \\
\midrule
A & Classic CNN & 15 & 0.596 & 0.255 & 0.732 & 0.564 & 21.26 & 0.554 \\
C & Lightweight CNN & 5 & 0.482 & 0.004 & 1.044 & 0.469 & 29.87 & 0.477 \\
D & Modern CNN & 8 & 0.585 & 0.183 & 0.764 & 0.597 & 19.32 & 0.590 \\
E & Plain Transformer & 4 & 0.466 & 0.002 & 2.884 & 0.193 & 33.11 & 0.329 \\
F & Hier. Transformer & 9 & 0.569 & 0.103 & 0.647 & 0.594 & 21.30 & 0.552 \\
G & CNN-Transf. Hybrid & 12 & 0.599 & 0.257 & 0.766 & 0.599 & 19.74 & 0.600 \\
H & MLP / attn-free & 9 & 0.466 & 0.114 & 0.985 & 0.490 & 32.46 & 0.496 \\
I & SSM / Mamba & 2 & 0.464 & 0.000 & 2.787 & 0.429 & 24.00 & 0.446 \\
\bottomrule
\end{tabular}
\end{table}

\begin{figure}[t]
    \centering
    \includegraphics[width=\columnwidth]{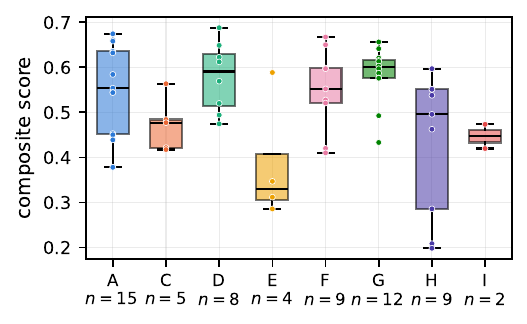}
    \caption{Composite score by group; each dot is one encoder, boxes show the
    median and quartiles. Within-group spread rivals the differences between
    group medians.}
    \label{fig:group}
\end{figure}

Table~\ref{tab:group} and Fig.~\ref{fig:group} aggregate encoders by group.
The strongest median group is the \BestGroup{}s (\BestGroupLetter{},
median score \BestGroupScore{}), narrowly ahead of the Modern CNNs; classical
CNNs and hierarchical transformers follow. The weakest by a wide margin is the
\WorstGroup{} group (\WorstGroupLetter{}, \WorstGroupScore{}): plain ViTs are
data-hungry, and without pretraining \NumPlainTDegen{} of the \NumPlainT{}
never acquire the boundaries that segmentation and thin-structure depth depend
on. The pattern inside the group is itself informative. The two members that
carry $\sim$100\,M parameters are the two worst, while the survivor,
\PlainTException{}, is also the smallest at \PlainTExceptionParams{}\,M and
reaches rank \PlainTExceptionRank{} --- so the group's poor showing is a
statement about the training regime, and if anything capacity makes it worse.

The more useful observation is about \emph{variance}. Within-group spread is
comparable to between-group spread: the classical-CNN group alone runs from the
second-best encoder overall down to nearly the worst. In other words,
\emph{which encoder} matters at least as much as \emph{which family} --- a group
label is a weak predictor of performance, and the representative chosen inside
the family carries most of the signal.

\subsection{Accuracy versus Cost}

\begin{table}[t]
\centering
\caption{Accuracy versus cost for the fifteen strongest encoders by composite
score. FPS and peak memory are measured end-to-end at $1920\times1080$, batch
1, mixed precision, on a single 16\,GB GPU. GFLOPs and memory cover both views
and both task branches.}
\label{tab:speed}
\scriptsize
\setlength{\tabcolsep}{3pt}
\renewcommand{\arraystretch}{0.95}
\begin{tabular}{@{}llrrrrrr@{}}
\toprule
Encoder & Grp & P$_{\text{e}}$(M) & GFLOPs & FPS & Mem(GB) & mIoU$\uparrow$ & $\delta_1\uparrow$ \\
\midrule
internimage\_t & D & 28.8 & 2213 & 3.7 & 3.5 & 0.650 & 0.725 \\
wrn\_40\_4 & A & 16.1 & 2128 & 9.4 & 3.1 & 0.649 & 0.699 \\
swin\_tiny\_w4 & F & 27.5 & 2184 & 4.0 & 3.8 & 0.633 & 0.701 \\
xception\_mobile\_order & A & 20.8 & 2267 & 7.8 & 3.5 & 0.634 & 0.687 \\
res2next50 & A & 22.6 & 2304 & 7.4 & 3.7 & 0.635 & 0.680 \\
edgenext\_xxs & G & 1.2 & 1414 & 9.2 & 2.9 & 0.626 & 0.686 \\
biformer\_tiny & F & 12.6 & 1777 & 4.5 & 3.3 & 0.597 & 0.702 \\
convnext\_atto & D & 3.4 & 1489 & 9.3 & 3.0 & 0.606 & 0.691 \\
coat\_lite\_tiny & G & 6.1 & 1699 & 6.0 & 3.1 & 0.615 & 0.667 \\
hrnet\_w18\_small & A & 3.9 & 1666 & 8.9 & 3.0 & 0.619 & 0.651 \\
inception\_v3\_small & A & 4.5 & 1703 & 8.1 & 3.2 & 0.615 & 0.655 \\
densenet121 & A & 7.0 & 2022 & 7.9 & 3.5 & 0.654 & 0.610 \\
unireplknet\_a & D & 4.0 & 1506 & 8.9 & 3.0 & 0.598 & 0.646 \\
ceit\_tiny\_nolca & G & 6.5 & 1677 & 3.2 & 3.0 & 0.629 & 0.612 \\
levit\_128s & G & 5.4 & 1450 & 4.5 & 2.9 & 0.622 & 0.608 \\
\bottomrule
\end{tabular}
\end{table}

\begin{figure}[t]
    \centering
    \includegraphics[width=\columnwidth]{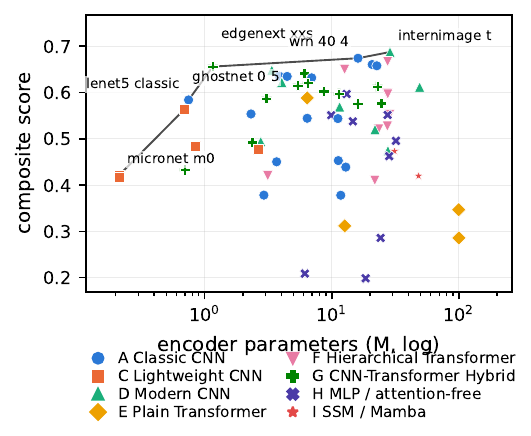}
    \caption{Composite score versus encoder parameters (log scale). The black
    curve is the Pareto front; compact convolutional and hybrid encoders reach
    it, and the largest models do not.}
    \label{fig:pareto}
\end{figure}

Table~\ref{tab:speed} and Fig.~\ref{fig:pareto} put quality against model size.
The headline is that parameter count does not buy accuracy on this task. The
tiny \SmallStrongEncoder{} (\SmallStrongParams{}\,M encoder parameters) sits on
the accuracy--size Pareto front, matching or beating encoders eighty times
larger, while the $\sim$100\,M transformers are both the slowest and the least
accurate. Among the leaders, the classical wide residual network and
\SmallStrongEncoder{} are also fast, whereas the top-ranked \BestEncoder{}
pays for its accuracy in latency. For a deployment budget, a compact modern or
classical convolutional encoder is the efficient choice; the large transformers
are dominated on every axis.

\subsection{Is the Ordering Real?}
\label{subsec:stability}

A ranking of \NumEncoders{} single runs invites the obvious objection that it
is noise. We separate two sources of that noise and can address only one of
them honestly.

\emph{Evaluation uncertainty} --- would a different test split reorder the
table? --- we can measure. The independent unit here is the \emph{scene}, not
the frame: the \NumFramesTest{} test frames are \NumScenesTest{} scenes seen
from 48 viewpoints each, and views of one tree are not independent samples. We
therefore cluster-bootstrap the \NumScenesTest{} test scenes
(\BootN{} resamples) and re-rank every encoder on each resample. The ordering
is stable: the re-ranked order correlates with the reported one at Spearman
$\rho=\BootRho{}$ (2.5th percentile \BootRhoLo{}), and \BestEncoder{} is
first in \BootTopOne{}\% of resamples. The reason the paired comparison is far
tighter than the marginal error bars would suggest is that every encoder is
scored on the identical frames, so scene difficulty is a common term that
cancels between encoders even though it dominates each encoder's absolute
score.

\emph{Training uncertainty} --- would a different seed reorder the table? ---
we cannot measure, because each encoder was trained once. \NumEncoders{}
encoders at multiple seeds was beyond the compute available, and this is the
sharpest limitation of the study: the analysis above bounds the noise from
\emph{which trees we tested on}, not from \emph{how the run happened to go}.
Differences of a few thousandths in Table~\ref{tab:main} should not be read as
ordering evidence; the effects we build conclusions on --- the collapse of
\NumDegenerate{} encoders, the gap between the top and bottom of the ranking,
and the seg--depth correlation --- are all far larger than that.

\subsection{Do the Two Tasks Prefer the Same Encoder?}

\begin{figure}[t]
    \centering
    \includegraphics[width=\columnwidth]{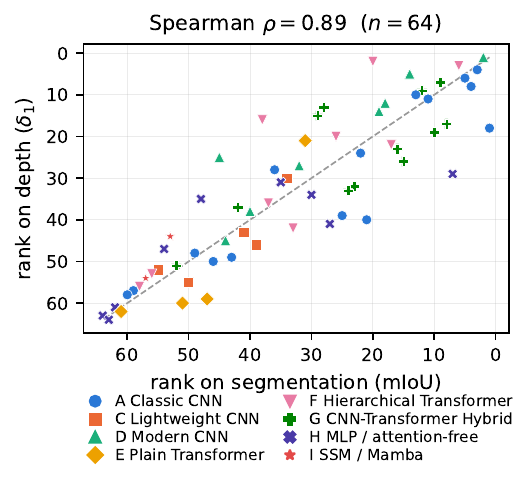}
    \caption{Rank on segmentation (mIoU) against rank on depth ($\delta_1$).
    Points cluster on the diagonal (top-left is best on both), giving a strong
    Spearman correlation and showing that the two tasks prefer the same
    encoders.}
    \label{fig:tradeoff}
\end{figure}

A shared encoder is only justified if segmentation and depth want the same
features. They do. Ranking the \RhoSegDepthN{} encoders by segmentation and by
depth gives a Spearman rank correlation of $\rho=\RhoSegDepth{}$
(Fig.~\ref{fig:tradeoff}): an encoder that is good for one task is almost always
good for the other. The practical consequence is that the shared-trunk design
costs almost nothing in this setting --- there is no meaningful subset of
encoders that trades segmentation for depth --- and that a single backbone
recommendation serves both heads.

\subsection{Qualitative Behaviour}
\label{subsec:qual}

\begin{figure*}[t]
    \centering
    \includegraphics[width=\textwidth]{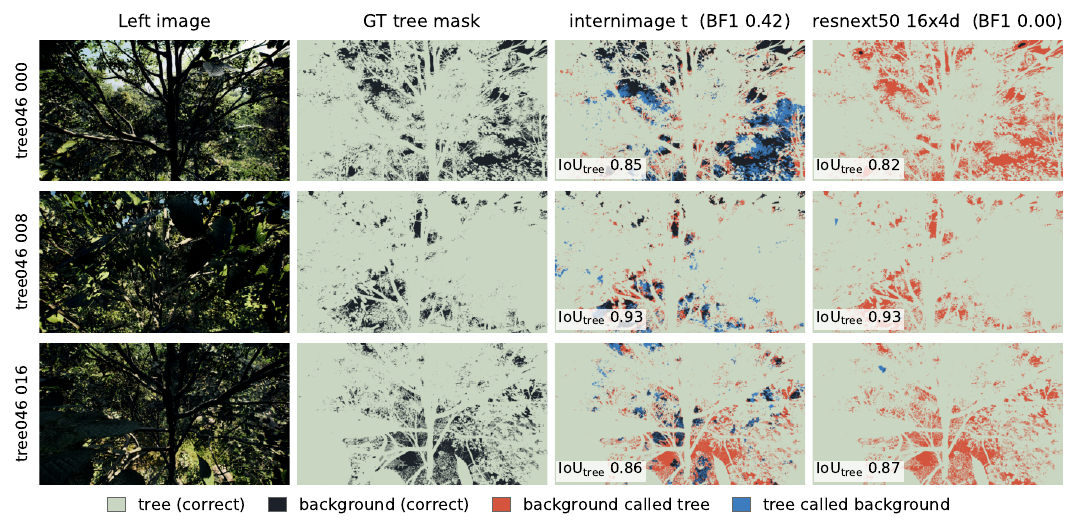}
    \caption{Qualitative segmentation on three held-out test scenes. Columns:
    left image, ground-truth tree mask, the best encoder (\BestEncoder{}) and a
    degenerate one (\DegenEncoder{}); the two prediction columns are coloured
    against the ground truth. The per-panel tree IoUs are within a few points of
    each other, so region IoU cannot separate the columns --- but
    \DegenEncoder{} has answered ``tree'' almost everywhere and floods every gap
    in the canopy with false positives (red), while \BestEncoder{} keeps the
    background between the branches open. Boundary F1, in the column headers,
    separates them at once.}
    \label{fig:qualseg}
\end{figure*}

\begin{figure*}[t]
    \centering
    \includegraphics[width=\textwidth]{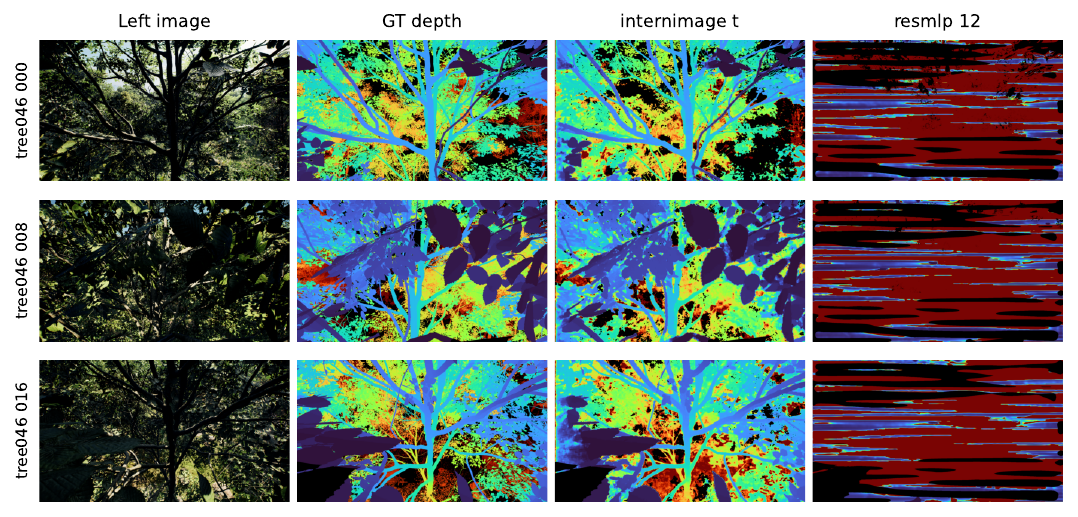}
    \caption{Qualitative depth on the same three scenes. Columns: left image,
    ground-truth depth, the best encoder (\BestEncoder{}), and the
    lowest-ranked encoder overall (\WeakEncoder{}). The strong encoder recovers
    the fine branch structure and tracks the ground truth; the weakest produces
    horizontally-streaked depth unrelated to the scene. Depth is drawn on each
    model's own predicted tree mask, which is why a segmentation failure is
    visible here as well.}
    \label{fig:qual}
\end{figure*}

The metric argument of Section~\ref{sec:eval} is not hypothetical: it is
visible in the predictions themselves (Fig.~\ref{fig:qualseg}).
\DegenEncoder{} --- a \DegenParams{}\,M classical residual network, precisely
the kind of backbone a project inherits on reputation --- reaches
\DegenTreeIoU{} tree IoU against \BestEncIoUTree{} for the best encoder, and on
that number alone would pass for competent. Its background
IoU is \DegenIoUBg{} and its boundary F1 is \DegenBFOne{}: it has learned to
answer ``tree'' and nothing else, and the silhouette the cutter needs is simply
absent. \BestEncoder{}, at \BestEncIoUBg{} background IoU and \BestEncBFOne{}
BF1, holds the gaps between the branches open instead. With
\NumDegenerate{} of \NumEncoders{} encoders in this state, the failure is the
common case rather than a curiosity.

The same encoder-driven gap appears in depth (Fig.~\ref{fig:qual}).
\BestEncoder{} tracks ground-truth depth closely and keeps thin twigs separated
from the background behind them, whereas \WeakEncoder{}, last of the
\NumEncoders{} by composite score (\WeakEncoderDelta{} $\delta_1$,
\WeakEncoderEPE{}\,px EPE), produces depth that bears little relation to the
scene. Because depth is consumed on the \emph{predicted} tree mask, a
segmentation failure propagates straight into the depth a downstream controller
would act on --- which is why the boundary and background terms, rather than
the near-saturated region IoU, drive the ranking.

\vspace{-0.3em}
\section{Discussion}

\textbf{How much does the encoder decide?} A great deal. Even after removing
the \NumDegenerate{} collapsed models, the \NumLive{} that remain span
\LiveMIoULo{}--\LiveMIoUHi{} mIoU and \LiveDeltaLo{}--\LiveDeltaHi{} depth
$\delta_1$ --- a factor of two on segmentation and effectively the whole usable
range on depth, from a change of encoder alone. The encoder is not a free
choice that a downstream head can compensate for; it sets the ceiling.

\textbf{From-scratch training reshapes the ranking.} The clearest single effect
is that data-hungry plain transformers collapse without pretraining ---
\NumPlainTDegen{} of \NumPlainT{} --- which is why group E sits at the bottom
despite two of its members being the largest models in the study. This is a
property of the training regime we chose in order to isolate architecture, and
it is the main reason our ordering should not be read as a universal backbone
ranking (see limitations).

\textbf{Architecture beats parameters.} The Pareto front is populated by compact
convolutional and hybrid encoders, and the largest models are dominated. On
thin-structure segmentation and depth, the inductive bias of the architecture
matters more than raw capacity --- the opposite of what a parameter-count
heuristic would suggest.

\textbf{One backbone is enough.} The strong seg--depth rank correlation means
the shared-encoder assumption is well founded here: there is no genuine
trade-off to manage, so a single recommendation --- a compact modern or
classical convolutional encoder --- serves both heads at once.

\vspace{-0.3em}
\section{Limitations}

Four limitations bound what these numbers support, in roughly decreasing order
of how much they should worry a reader.

\textbf{Single seed.} Every encoder was trained once. Section~\ref{subsec:stability}
bounds the noise from the choice of test scenes but says nothing about run-to-run
variance, so small differences in Table~\ref{tab:main} carry no weight; only the
large effects should be read as findings.

\textbf{Synthetic data, and only \NumScenesTest{} test scenes.} All conclusions
are on rendered forest scenes. Exact geometric labels are what make
thin-structure evaluation possible at all --- no hand annotation resolves a
two-pixel twig --- but rendered canopies lack the sensor noise, motion blur and
illumination of field imagery, and transfer is untested here. The test split is
also narrow: \NumScenesTest{} trees.

\textbf{No pretraining.} Training from scratch isolates architecture but
removes an advantage that some families, plain transformers in particular,
benefit from more than others. Our ordering should be read as a from-scratch
ordering, not a universal backbone ranking; establishing whether pretraining
reorders it is the obvious next experiment.

\textbf{One representative per family, unequal budgets.} Within-family variance
is not measured and a different preset could move a family. Four families are
represented only at $\sim$100\,M because they publish nothing smaller, though
their weak results cut against capacity rather than for it. \NumEncoders{}
encoders completed; the remainder exceeded the memory budget or had not
finished.

\vspace{-0.3em}
\section{Conclusion}

We asked how much the encoder decides on joint tree segmentation and stereo
depth, and answered it by swapping only the encoder --- \NumEncoders{} of them,
one per architecture family, trained from scratch --- through a
fixed segmentation branch, a fixed stereo branch and an evaluation built to
resist the label-everything-tree shortcut. The encoder decides a great deal:
compact convolutional and hybrid backbones lead (\BestEncoder{} best overall),
\NumPlainTDegen{} of \NumPlainT{} plain transformers collapse without
pretraining, parameter count fails to
predict quality, and segmentation and depth agree strongly enough
($\rho=\RhoSegDepth{}$) that one shared backbone serves both. For thin-structure
perception in forestry robotics, the practical recommendation is to choose a
compact modern or classical convolutional encoder by measurement on the target
task rather than by inheriting whichever backbone is strong on ImageNet.

\vspace{-0.2em}
\section*{Reproducibility and Availability}

The sweep is one resumable command that trains and evaluates each encoder in an
isolated subprocess and records per-encoder failures rather than aborting.
Every table, figure and in-text number above --- including the bootstrap of
Section~\ref{subsec:stability} and the masks in Fig.~\ref{fig:qualseg}, which
are recovered from the stored predictions and checked against each run's own
per-frame record --- is emitted from the result files by a single script;
nothing is transcribed by hand. Code, the tree50 renderer and the trained
checkpoints will be released with the paper.

    \balance
    \bibliographystyle{IEEEtran}

\begin{thebibliography}{99}
        \bibitem{he2016resnet} K. He, X. Zhang, S. Ren, and J. Sun, ``Deep residual
            learning for image recognition,'' in \emph{CVPR}, 2016.
        \bibitem{simonyan2015vgg} K. Simonyan and A. Zisserman, ``Very deep
            convolutional networks for large-scale image recognition,'' in
            \emph{ICLR}, 2015.
        \bibitem{huang2017densenet} G. Huang, Z. Liu, L. van der Maaten, and K. Q.
            Weinberger, ``Densely connected convolutional networks,'' in
            \emph{CVPR}, 2017.
        \bibitem{xie2017resnext} S. Xie, R. Girshick, P. Doll\'ar, Z. Tu, and K. He,
            ``Aggregated residual transformations for deep neural networks,'' in
            \emph{CVPR}, 2017.
        \bibitem{howard2019mobilenetv3} A. Howard \emph{et al.}, ``Searching for
            MobileNetV3,'' in \emph{ICCV}, 2019.
        \bibitem{ma2018shufflenetv2} N. Ma, X. Zhang, H.-T. Zheng, and J. Sun,
            ``ShuffleNet V2: Practical guidelines for efficient CNN architecture
            design,'' in \emph{ECCV}, 2018.
        \bibitem{ding2021repvgg} X. Ding \emph{et al.}, ``RepVGG: Making VGG-style
            ConvNets great again,'' in \emph{CVPR}, 2021.
        \bibitem{dosovitskiy2021vit} A. Dosovitskiy \emph{et al.}, ``An image is
            worth 16x16 words: Transformers for image recognition at scale,'' in
            \emph{ICLR}, 2021.
        \bibitem{touvron2021deit} H. Touvron \emph{et al.}, ``Training
            data-efficient image transformers and distillation through attention,''
            in \emph{ICML}, 2021.
        \bibitem{liu2021swin} Z. Liu \emph{et al.}, ``Swin Transformer: Hierarchical
            vision transformer using shifted windows,'' in \emph{ICCV}, 2021.
        \bibitem{wang2021pvt} W. Wang \emph{et al.}, ``Pyramid Vision Transformer: A
            versatile backbone for dense prediction without convolutions,'' in
            \emph{ICCV}, 2021.
        \bibitem{liu2022convnext} Z. Liu, H. Mao, C.-Y. Wu, C. Feichtenhofer, T.
            Darrell, and S. Xie, ``A ConvNet for the 2020s,'' in \emph{CVPR}, 2022.
        \bibitem{ding2022replknet} X. Ding, X. Zhang, J. Han, and G. Ding, ``Scaling
            up your kernels to 31x31: Revisiting large kernel design in CNNs,'' in
            \emph{CVPR}, 2022.
        \bibitem{dai2021coatnet} Z. Dai, H. Liu, Q. V. Le, and M. Tan, ``CoAtNet:
            Marrying convolution and attention for all data sizes,'' in
            \emph{NeurIPS}, 2021.
        \bibitem{yu2022metaformer} W. Yu \emph{et al.}, ``MetaFormer is actually
            what you need for vision,'' in \emph{CVPR}, 2022.
        \bibitem{tolstikhin2021mlpmixer} I. Tolstikhin \emph{et al.}, ``MLP-Mixer:
            An all-MLP architecture for vision,'' in \emph{NeurIPS}, 2021.
        \bibitem{trockman2023convmixer} A. Trockman and J. Z. Kolter, ``Patches are
            all you need?,'' \emph{TMLR}, 2023.
        \bibitem{liu2024vmamba} Y. Liu \emph{et al.}, ``VMamba: Visual state space
            model,'' in \emph{NeurIPS}, 2024.
        \bibitem{zhu2024vim} L. Zhu \emph{et al.}, ``Vision Mamba: Efficient visual
            representation learning with bidirectional state space model,'' in
            \emph{ICML}, 2024.
        \bibitem{caruana1997multitask} R. Caruana, ``Multitask learning,''
            \emph{Machine Learning}, vol. 28, no. 1, pp. 41--75, 1997.
        \bibitem{kendall2018multitask} A. Kendall, Y. Gal, and R. Cipolla,
            ``Multi-task learning using uncertainty to weigh losses for scene
            geometry and semantics,'' in \emph{CVPR}, 2018.
        \bibitem{misra2016crossstitch} I. Misra, A. Shrivastava, A. Gupta, and M.
            Hebert, ``Cross-stitch networks for multi-task learning,'' in
            \emph{CVPR}, 2016.
        \bibitem{vandenhende2021mtlsurvey} S. Vandenhende \emph{et al.},
            ``Multi-task learning for dense prediction tasks: A survey,''
            \emph{IEEE TPAMI}, 2021.
        \bibitem{kendall2017gcnet} A. Kendall \emph{et al.}, ``End-to-end learning
            of geometry and context for deep stereo regression,'' in
            \emph{ICCV}, 2017.
        \bibitem{chang2018psmnet} J.-R. Chang and Y.-S. Chen, ``Pyramid stereo
            matching network,'' in \emph{CVPR}, 2018.
        \bibitem{guo2019gwcnet} X. Guo \emph{et al.}, ``Group-wise correlation
            stereo network,'' in \emph{CVPR}, 2019.
        \bibitem{ronneberger2015unet} O. Ronneberger, P. Fischer, and T. Brox,
            ``U-Net: Convolutional networks for biomedical image segmentation,''
            in \emph{MICCAI}, 2015.
        \bibitem{bac2014harvesting} C. W. Bac \emph{et al.}, ``Harvesting robots
            for high-value crops: State-of-the-art review and challenges
            ahead,'' \emph{J. Field Robotics}, vol. 31, no. 6, 2014.
        \bibitem{zahid2021pruner} A. Zahid \emph{et al.}, ``Technological
            advancements towards developing a robotic pruner for apple trees: A
            review,'' \emph{Comput. Electron. Agric.}, vol. 189, 2021.
        \bibitem{you2022pruning} A. You \emph{et al.}, ``An autonomous robot for
            pruning modern, planar fruit trees,'' \emph{arXiv:2206.07201}, 2022.
        \bibitem{lahera2024forestry} P. La Hera \emph{et al.}, ``Exploring the
            feasibility of autonomous forestry operations,'' \emph{J. Field
            Robotics}, vol. 41, 2024.
        \bibitem{madaan2017wire} R. Madaan, D. Maturana, and S. Scherer, ``Wire
            detection using synthetic data and dilated convolutional networks
            for UAVs,'' in \emph{IROS}, 2017.
        \bibitem{barth2018synthesis} R. Barth \emph{et al.}, ``Data synthesis
            methods for semantic segmentation in agriculture: A
            \emph{Capsicum annuum} dataset,'' \emph{Comput. Electron. Agric.},
            vol. 144, 2018.
    \end{thebibliography}
    
\end{document}